\pdfoutput=1
\documentclass[11pt]{article}

\usepackage{emnlp2021}

\usepackage{times}
\usepackage{latexsym}

\usepackage[T1]{fontenc}
\usepackage[utf8]{inputenc}

\usepackage{microtype}

\usepackage{multirow}
\usepackage{graphicx}

\newenvironment{myquote}%
  {\list{}{\leftmargin=0.15in\rightmargin=0.15in}\item[]}%
  {\endlist}

\title{Probing Language Models for Understanding of Temporal Expressions}

\author{Shivin Thukral
  \and
  Kunal Kukreja
  \and 
  Christian Kavouras \\
  Department of Linguistics \\
  University of Washington \\
  \texttt{\{shivin7, kkukreja, cdkavour\}@uw.edu} \\}

\begin{document}
\maketitle

\begin{abstract}
We present three Natural Language Inference (NLI) challenge sets that can evaluate NLI models on their understanding of temporal expressions. More specifically, we probe these models for three temporal properties: (a) the order between points in time, (b) the duration between two points in time, (c) the relation between the magnitude of times specified in different units. We find that although large language models fine-tuned on MNLI have some basic perception of the order between points in time, at large, these models do not have a thorough understanding of the relation between temporal expressions.
\end{abstract}

\section{Introduction}

While contextualized embeddings obtained from recent transformer-based models such as BERT \citep{devlin-etal-2019-bert} have proven to contain a lot of semantic and syntactic information about the tokens they encode, recent studies have shown that there are still gaps in their understanding \citep{rogers-etal-2020-primer}. On the semantic side, for instance, BERT struggles with representations of numbers \citep{wallace-etal-2019-nlp} and cannot reason based on its world knowledge \citep{rogers-etal-2020-primer}. Work in NLI has also developed challenge sets showing that the reported performance of these language models on various tasks can be exaggerated \citep{mccoy-etal-2019-right}, and they rely on lexical cues in the dataset instead of actual language comprehension.

Our work explores the grasp of such models on the relation between temporal expressions. Temporal expressions, or time expressions, in text are a sequence of tokens that denote time, such as a point in time (6 May 1980, Monday, 12 PM) or duration (7 minutes, 5 years, 2 months). More specifically, we try to determine whether these models capture the ordering and duration relationships between different points in time. We also analyze if these models can reason about durations specified in different units. Recognition of temporal expressions has had applications in timeline construction \citep{do-etal-2012-joint, leeuwenberg-moens-2018-temporal} and clinical analysis \citep{bethard-etal-2015-semeval} previously, and can be beneficial for dialogue assistants in scheduling reminders and meetings, which shapes our motivation behind conducting such an analysis. We evaluate these models on the above temporal properties by presenting three NLI challenge sets.

Our experiments demonstrate that language models such as RoBERTa \citep{liu2019roberta} and DeBERTa \citep{he2021deberta} fine-tuned on existing large NLI datasets are unable to completely reason about the ordering and duration between temporal expressions. We further analyze the examples and find that while these models recognize whether a point in time lies within an interval, they cannot capture other relations between time instances and durations\footnotemark[1].

\footnotetext[1]{Code and data available on \href{https://github.com/kunalkukreja21/temporal-expressions-evaluation-lm}{GitHub}}

\section{Related Work}

Much work has been done on the extraction of events, temporal expressions \citep{timeexpressionrecognition, Ding_Gao_Shi_Qu_2019}, and the temporal relations between the two. TimeBank \citep{pustejovsky2003timebank} was one of the first annotated corpora for this task. It utilized the TimeML \citep{pustejovsky2003timeml} standard for annotation. TempEval-1 \citep{verhagen-etal-2007-semeval}, TempEval-2 \citep{verhagen-etal-2010-semeval}, and TempEval-3 \citep{uzzaman-etal-2013-semeval} are shared tasks created for evaluating models on various temporal properties, and most methods used were traditional rule-based \citep{strotgen-gertz-2010-heideltime, ning-etal-2018-cogcomptime} or grammar-based \citep{lee-etal-2014-context} solutions.

Various corpora have been developed that test for different temporal properties. \citet{vashishtha-etal-2019-fine} map events to their fine-grained duration, and event pairs to their relative timelines. \citet{naik-etal-2019-tddiscourse} create additional annotations in the existing TimeBank-Dense corpus \citep{cassidy-etal-2014-annotation} for discourse-level temporal ordering. \citet{ning2020torque} create a reading comprehension dataset that tests for temporal ordering. \citet{zhou-etal-2019-going} test for various temporal commonsense properties using a multiple-choice question-answering dataset. \citet{vashishtha-etal-2020-temporal} recast existing temporal datasets into NLI format to test for temporal ordering and duration. 

Our goal is to create similar datasets to probe for a semantic understanding of temporal expressions in pre-trained language models. We create these datasets in an NLI format and use them to evaluate NLI models trained on MNLI \citep{williams-etal-2018-broad}, which is a generic NLI dataset. We choose MNLI because it is large and diverse. The dataset contains time terms in 36\% of the development instances, including examples containing temporal expressions like months and days of the week. We investigate whether these examples are sufficient for a general perception of temporal expressions. 

To our knowledge, there has been little work investigating the implicit understanding of time expressions in pre-trained large language models. The most similar work to ours is \citet{vashishtha-etal-2020-temporal}. They produce five NLI datasets recast from existing temporal reasoning corpora and test NLI models for event duration (how long an event lasts) and event ordering (how events are temporally arranged). However, there are some key differences:
\begin{itemize}
    \item Our focus is to investigate the temporal properties of ordering and duration for explicit \textit{time expressions}, and not for \textit{events} in a sentence.
    \item We analyze whether language models can reason about more fine-grained duration (e.g., whether an event takes exactly 5 hours) where as they analyze reasoning about more coarse-grained duration (e.g., whether an event takes place in the order of hours or days).
    \item We also investigate whether language models can figure out commonplace conversions among adjacent units of time.
    \item We introduce numerous variations in our data creation process about how the time expressions are inserted and draw conclusions from how these variations affect performance.
\end{itemize}

\section{Dataset Creation}

We construct three NLI datasets that aim to test different relations between temporal expressions. The datasets use templates from a manually curated list of 71 events, labeled with their \textit{temporal occurrence} (when the event is likely to occur) and \textit{temporal duration} (how long the event is expected to last) values. For instance:

\begin{myquote}
    \textit{Template}: I went to Paris \\
    \textit{Occurrence}: day, month, year \\
    \textbullet \hspace{0.3mm} I went to Paris on Monday. \\
    \textbullet \hspace{0.3mm} I went to Paris in March. \\
    \textbullet \hspace{0.3mm} I went to Paris in 2010. \\
    \textit{Duration}: hours, days \\
    \textbullet \hspace{0.3mm} I visited Paris from 10 AM to 9 PM. \\
    \textbullet \hspace{0.3mm} I visited Paris from Mon to Wed.
\end{myquote}

Each temporal unit corresponds to some list(s) spanning different magnitudes of time (Table \ref{tab:Temporal Expressions Lists}), which are used during NLI pair creation.

\begin{table}
    \centering
    \begin{tabular}{ll}
    \hline
     \textbf{List Type} & \textbf{List Range} \\
      \hline
      hour (12 hr) & \textit{12 AM, ..., 11 PM} \\
      hour (24 hr) & \textit{00:00, 01:00, ..., 23:00} \\
      weekday & \textit{Sunday, ..., Saturday} \\
      month-day & \textit{1st, 2nd, ..., 28th} \\
      month (full name) & \textit{January, ..., December} \\
      month (abbreviated) & \textit{Jan, Feb, ..., Dec} \\
      year & \textit{1900, 1901, ..., 2000} \\
      \hline
    \end{tabular}
    \caption{Different lists of temporal expressions}
    \label{tab:Temporal Expressions Lists}
\end{table}

\subsection{Set I: Temp-Order}

We create this NLI challenge set to test whether language models recognize the relationship of ordering between two distinct temporal expressions. We frame this in the NLI format by having the premise mention an event occur at a particular time instance, while the hypothesis mentions the same event but occurring at a different time instance:

\begin{myquote}
    \textit{Premise} : They got married \textbf{in March}. \\
    \textit{Hypothesis} : They got married \textbf{before July}. \\  
    \textit{Label} : Entailment
\end{myquote}

\begin{table*}
    \centering
    \begin{tabular}{lllc}
    \hline
     & \textbf{Premise} & \textbf{Hypothesis} & \textbf{Label} \\
       \hline
      a) & He left his job \textit{at 12 PM}. & He left his job \textit{before 5 PM}. & E \\
      b) & \textit{At 12 PM}, he left his job. & \textit{Before 5 PM}, he left his job. & E \\
      c) & He will leave his job \textit{at 12 PM}. & He will leave his job \textit{before 5 PM}. & E \\
      d) & He left his job \textit{after 12 PM}. & He left his job \textit{after 9 AM}. & E \\
      e) & He left his job \textit{after 12 PM}. & He left his job \textit{before 5 PM}. & N \\
      f) & He left his job \textit{after 12 PM}. & He left his job \textit{before 9 AM}. & C \\
      g) & He left his job \textit{at 12 PM}. & He left his job \textit{before 17:00}. & E \\
      h) & He left his job \textit{in February}. & He left his job \textit{after Apr}. & C \\
      i) & He left his job \textit{in October 2011}. & He left his job \textit{after Jan 2011}. & E \\
      j) & He left his job \textit{on 21st Sep 2013}. & He left his job \textit{before 23rd Sep 2012}. & C \\
      \hline
    \end{tabular}
    \caption{Variations in NLI pairs for ordering of temporal expressions (E$\,\to\,$entailment, C$\,\to\,$contradiction, N$\,\to\,$neutral). \textit{a)} is the basic construction; \textit{b), c)} is with the variation in event template; \textit{d), e), f)} are when premise uses a relative preposition to allow the event to happen in a time interval; \textit{g), h)} are examples of choosing time instances from two different lists; \textit{i), j)} are generation of more specific dates using months and month-days with years.}
    \label{tab:Variations in Temporal Ordering}
\end{table*}

We start constructing a basic NLI pair by choosing a sentence template from the list of events. Based on the \textit{temporal occurrence} label of the event, one of the lists from Table \ref{tab:Temporal Expressions Lists} is chosen, and two time instances are sampled from that list with replacement. For the premise, the first time instance is attached so that the event happens precisely at this time instance. For the hypothesis, we randomly choose a \textit{relative ordering} between \textit{`before'} and \textit{`after'} and attach it to the template event and the second instance. Since the premise claims that the event occurs at an exact point in time while the hypothesis claims that the event happens in a specific time interval, the premise time instance either lies inside the hypothesis time interval or it does not, generating the labels of \textit{entailment} or \textit{contradiction} correspondingly. 

During label generation, we have assumed that both time instances lie in the same cycle (e.g., two \textit{weekdays} lie in the same week). However, for cases where the two time instances are close across consecutive cycles, the automated label generated this way might be considered conventionally wrong:

\begin{myquote}
    \textit{Premise}: The concert starts \textbf{at 2 AM}. \\
    \textit{Hypothesis}: The concert starts \textbf{before 11 PM}. \\  
    \textit{Label}: Entailment
\end{myquote}

To reduce the number of such edge cases in the dataset, we do not allow sampling of time instances that are more than half the length of the list far apart (e.g., for within a day, the distance between two \textit{hours} will be at most 12).

We also introduce some variations in the sentence generation process to analyze the sensitivity of the models. Firstly, we tweak the event template by changing its position in the sentence (Table \ref{tab:Variations in Temporal Ordering} \textit{b}) and by switching it to future tense (Table \ref{tab:Variations in Temporal Ordering} \textit{c}). Secondly, we allow the premise event to also occur over an interval of time rather than a point in time (Table \ref{tab:Variations in Temporal Ordering} \textit{d, e, f}). To generate labels for these cases, the criteria we follow is that the pair is an \textit{entailment} if the premise time interval is completely included in the hypothesis time interval (\textit{temporal inclusion}), a \textit{contradiction} if there is no overlap between the two (\textit{temporal precedence}), and \textit{neutral} otherwise. Moreover, we allow the premise and hypothesis to sample points in time from different lists when possible (Table \ref{tab:Variations in Temporal Ordering} \textit{g, h}). We also generate more specific dates by combining \textit{months} and \textit{month-days} with \textit{years} (Table \ref{tab:Variations in Temporal Ordering} \textit{i, j}) to see if the language models are still able to reason about the difference in their ordering. 

We construct separate train and test datasets, using 53 templates for the train split and 18 templates for the test split. We have 11 different ways of choosing the two time instances: seven ways of choosing both from the same list (Table \ref{tab:Temporal Expressions Lists}) and four ways of choosing from different lists (Table \ref{tab:Variations in Temporal Ordering} \textit{g-j}). We choose the two time instances for each template based on its \textit{temporal occurrence} label and run it for five iterations, which results in a train dataset of 16,980 instances and test dataset of 6,140 instances, with the distribution of labels being: 40\% \textit{contradiction}, 35\% \textit{entailment}, 25\% \textit{neutral}.

\subsection{Set II: Temp-Duration}

The motivation behind this dataset is to test whether language models can reason about fine-grained temporal durations. We frame this in an NLI format by having the premise mention an event occurring between two points in time, while the hypothesis mentions the same event having occurred for a given duration:

\begin{table*}
    \centering
    \begin{tabular}{lp{0.4\linewidth}lc}
    \hline
    & \textbf{Premise} & \textbf{Hypothesis} & \textbf{Label} \\
       \hline
      a) & The meeting lasted \textit{from 12 PM to 5 PM}. & The meeting lasted \textit{for 5 hours}. & E \\
      b) & The meeting lasted \textit{from 12 PM to 5 PM}. & The meeting lasted \textit{for 50 hours}. & C \\
      c) & The meeting lasted \textit{from 12 PM to 5 PM}. & The meeting lasted \textit{for less than 5 hours}. & C \\
      d) & The meeting lasted \textit{from 12 PM to 5 PM}. & The meeting lasted \textit{for less than 6 hours}. & E \\
      e) & The meeting \textit{began at 12 PM} and \textit{lasted until 5 PM}. & The meeting lasted \textit{for 5 hours}. & E \\
      f) & The meeting lasted \textit{from 9 PM to 3 AM}. & The meeting lasted \textit{for 6 hours}. & E \\
      g) & The meeting lasted \textit{from 12 PM to 17:00}. & The meeting lasted \textit{for 5 hours}. & E \\
      h) & The spring quarter lasts \textit{from Mar to June}. & The spring quarter lasts \textit{for 3 months}. & E \\
      i) & The war lasted \textit{from July 1914 to Nov 1918}. & The war lasted \textit{for 4 years 4 months}. & E \\
      j) & The war lasted \textit{from July 1914 to Nov 1918}. & The war lasted \textit{for 52 months}. & E \\
      \hline
    \end{tabular}
    \caption{Variations in NLI pairs for duration calculation (E$\,\to\,$entailment, C$\,\to\,$contradiction). \textit{a) - d)} are a few examples from the 6 basic pairs; \textit{e)} is with a changed premise structure; \textit{f)} is when the hypothesis time instance crosses over to the next cycle; \textit{g), h)} are examples of choosing time instances from two different lists; \textit{i), j)} are generation of specific dates using months and years in two different formats.}
    \label{tab:Variations in Temporal Duration}
\end{table*}

\begin{myquote}
    \textit{Premise} : The war lasted \textbf{from 1939 to 1945}. \\
    \textit{Hypothesis} : The war lasted \textbf{for 6 years}. \\  
    \textit{Label} : Entailment
\end{myquote}

We begin forming a basic NLI pair by choosing a sentence template. Based on the event's \textit{temporal duration} label, a list from Table \ref{tab:Temporal Expressions Lists} is selected, and two time instances are randomly sampled without replacement. The smaller instance is mentioned in the premise as the event start time and the other instance as the event end time. We construct multiple hypotheses for the same premise. First, we calculate the \textit{gold duration (GOLD)} by finding the difference between the two instances, assuming both the instances are part of the same cycle. Then, the hypothesis mentions the event to have occurred in two different settings \textit{(equal to, less than)} for three different durations \textit{(GOLD, GOLD+1, GOLD*10)}, generating a total of six hypotheses (Table \ref{tab:Variations in Temporal Duration} \textit{a-d} are a few examples). We do this to test whether the NLI models can reason for the claimed duration's validity only when they are very distant \textit{(GOLD*10)} or also very close \textit{(GOLD+1)} to the gold duration. Generation of true labels for the pairs is automated, producing an \textit{entailment} or \textit{contradiction} depending on whether the gold duration falls in the duration range specified by the hypothesis.

We again introduce two variations in the dataset creation process. First, we change the wording of the premise sentence (Table \ref{tab:Variations in Temporal Duration} \textit{e}). Secondly, while sampling the time instances, we force the ending instance to be picked such that it falls before the starting instance in the list, which implies that the event crossed over to the next cycle. In such cases, the calculation of the gold duration is slightly different (Table \ref{tab:Variations in Temporal Duration} \textit{f}). We perform this next cycle calculation for all lists in Table \ref{tab:Temporal Expressions Lists} except \textit{month-days} (because gold calculation without specifying the exact month is ambiguous) and \textit{years} (because the list is acyclic). We also allow a similar blend of temporal expressions, like in \textit{Temp-Order} set, combining the two \textit{hours} (Table \ref{tab:Variations in Temporal Duration} \textit{g}) and \textit{months} (Table \ref{tab:Variations in Temporal Duration} \textit{h}) lists. We construct specific dates by including \textit{months} and \textit{years} and allow the duration to be mentioned in a year-month format (Table \ref{tab:Variations in Temporal Duration} \textit{i}) or a months-only format (Table \ref{tab:Variations in Temporal Duration} \textit{j}). 

We create separate train and test datasets using the same split of 53 and 18 templates as before. For each template, time instances are chosen based on their \textit{temporal duration} label, along with the variations as mentioned above applied (Table \ref{tab:Variations in Temporal Duration} \textit{e-j}). Running each template for five iterations produces a train dataset of 13,500 instances and test dataset of 3,540 instances, with the label distribution: 50\% \textit{entailment} and 50\% \textit{contradiction}.

\begin{table*}
    \centering
    \begin{tabular}{lllc}
    \hline
    & \textbf{Premise} & \textbf{Hypothesis} & \textbf{Label} \\
       \hline
      a) & The store will close \textit{in 2 hours}. & The store will close \textit{before 40 minutes}. & C \\
      b) & \textit{In 2 hours}, the store will close. & The store will close \textit{after 84 minutes}. & E \\
      c) & The store will close \textit{in 2 days}. & \textit{After 34 hours}, the store will close. & E \\
      d) & \textit{After 4 days}, the store will close. & The store will close \textit{before 38 hours}. & C \\
      e) & The store will close \textit{before 4 days}. & \textit{Before 174 hours}, the store will close. & E \\
      f) & The store will close \textit{before 6 hours}. & The store will close \textit{after 77 minutes}. & N \\
      g) & \textit{After 3 hours}, the store will close. & The store will close \textit{after 409 minutes}. & N \\
      \hline
    \end{tabular}
    \caption{Variations in NLI pairs for cross-unit duration comparison (E$\,\to\,$entailment, C$\,\to\,$contradiction, N$\,\to\,$neutral). \textit{a)} is the basic pair; \textit{b), c)} are variations of basic pair with template position changed; \textit{d) - g)} are variations in which the premise event occurs over a range of time.}
    \label{tab:Variations in Cross-Magnitude}
\end{table*}

\subsection{Set III: Cross-Unit Duration}

The motivation behind the creation of the \textit{Cross-Unit Duration} set is to test whether language models understand the conversion relationship between magnitudes specified in different units of time; for instance, if models are able to interpret that 5 hours are less than 350 minutes but more than 250. Moreover, we investigate if these models are better at certain kinds of conversions. We frame this task in an NLI format in a similar manner to the \textit{Temp-Order} set. In the premise, we mention a future event that will occur after a given duration \textit{(T1)}, while in the hypothesis we mention the same future event to occur before or after a different duration \textit{(T2)}. Apart from varying magnitudes, \textit{T1} and \textit{T2} are also specified in different but adjacent units of time. More specifically, \textit{T1} is specified in the higher adjacent unit of time, i.e., if \textit{T2} is specified in minutes, then \textit{T1} will be specified in hours. Since the premise mentions an event occurring at a future point in time while the hypothesis mention an event occurring over a time interval bounded on just one side, the premise event either lies in the interval or not, leading to the labels \textit{entailment} and \textit{contradiction} respectively. We tried multiple variations similar to \textit{Temp-Order} set, like changing the position of the template in the premise/hypothesis (Table \ref{tab:Variations in Cross-Magnitude} \textit{b, c}) and making the event in the premise also occur over a future interval (Table \ref{tab:Variations in Cross-Magnitude} \textit{d-g}). The labeling procedure of the second variation is again similar to \textit{Temp-Order} set.

To create the challenge set, we first pick a template and look at the list of its \textit{temporal occurrence} values. We then iterate over all adjacent values in this list, e.g., seconds-minutes, hours-days, months-years. For each pair, we iterate over a manually created list of magnitudes for the higher unit of time \textit{(T1)} for the premise. We then pick a magnitude in the smaller unit of time \textit{(T2)} which is either higher or lower than \textit{T1}. \textit{T2}'s value is generated randomly, but a \textit{difference range} parameter controls its absolute difference with \textit{T1}'s value. For each fixed template, fixed duration unit pair, and fixed magnitude of \textit{T1}, we generate twelve different premise-hypothesis pairs, four in which the premise occurs at a future point in time, and eight in which it occurs over a future interval of time. Using a \textit{difference range} parameter of 5, we create a training set of 42,240 rows and a test set of 15,840 rows. Due to the challenge set creation procedure, the resultant dataset is naturally balanced with the same number of samples for each of the three labels. 

\section{Experimental Setup}

We evaluate three different NLI models on each of our challenge sets. The first model is a pre-trained RoBERTa-large model fine-tuned on the MNLI corpus, which reports 90.8\% accuracy on the MNLI-matched task. The second model is Microsoft's DeBERTa-large model fine-tuned on the MNLI corpus, which reports 91.9\% accuracy on the MNLI-matched task. Both these models are trained on all three labels: \textit{entailment}, \textit{contradiction}, and \textit{neutral}.

The third model comes from \citet{vashishtha-etal-2020-temporal}, which is a RoBERTa-large model fine-tuned on their temporal NLI datasets. For \textit{Temp-Order} set, we evaluate their model trained on the \textit{UDS-NLI (order)} corpus as it explored ordering relations between events, and we wanted to analyze if any of that knowledge transfers over for determining the ordering between temporal expressions. Similarly, for \textit{Temp-Duration} and \textit{Cross-Unit Duration} sets, we evaluate their model trained on the \textit{UDS-NLI (duration)} corpus, which explored more coarse-grained duration of events. In contrast to the models fine-tuned on MNLI, these models are only trained on binary classification - producing \textit{`entailed'} for the \textit{entailment} label and \textit{`not-entailed'} for the \textit{contradiction} and \textit{neutral} labels. We have a separate majority baseline corresponding to these models.

\begin{table*}
\centering
    \begin{tabular}{cccc}
    \hline
      \textbf{Model} & \textbf{Method} & \textbf{Accuracy} & \textbf{F1 Score} \\
      \hline
      \multirow{2}{*}{\textbf{Majority}} & Ternary Classification & 40.29 & 23.14 \\
      & Binary Classification & 65.19 & 51.46 \\
      \hline
      \multirow{3}{*}{\textbf{RoBERTa (MNLI)}} & Direct Evaluation & 52.75 & 45.36 \\
      & Hypothesis Only Training & 40.29 $\pm$ 0 & 23.14 $\pm$ 0 \\
      & Train and Evaluate & 99.81 $\pm$ 0.03 & 99.81 $\pm$ 0.03 \\
      \hline
      \multirow{3}{*}{\textbf{DeBERTa (MNLI)}} & Direct Evaluation & 51.57 & 44.29 \\
      & Hypothesis Only Training & 40.29 $\pm$ 0 & 23.14 $\pm$ 0 \\
      & Train and Evaluate & 99.76 $\pm$ 0.04 & 99.76 $\pm$ 0.04 \\
      \hline
      \multirow{1}{*}{\textbf{UDS-NLI (order)}} & Direct Evaluation & 56.36 & 57.20 \\
      \hline
    \end{tabular}
    \caption{Evaluating \textit{Temp-Order} set on NLI models}
    \label{tab:Results CS1}
\end{table*}

We report performances of all datasets under three different settings:
\begin{enumerate}
    \item \textbf{Direct Evaluation}: Evaluating the pre-trained NLI models directly on the test splits of our challenge sets.
    \item \textbf{Train and Evaluate}: Fine-tuning the NLI models with the train splits and reporting performances on the test splits. We report this to recognize the complexity of the synthetic datasets, and the ceiling performances that various NLI models can achieve on them.
    \item \textbf{Hypothesis Only Training}: Fine-tuning the NLI models in a hypothesis-only setting \citep{poliak-etal-2018-hypothesis} with the train splits, and reporting performances on the test splits. We report this as a control for the results achieved in the \textit{Train and Evaluate} setting.
\end{enumerate}

We do not train the models fine-tuned on UDS-NLI corpora, and only report their performance under the \textit{Direct Evaluation} setting, as the architecture of those models is similar to that of the pre-trained RoBERTa-MNLI model and we hypothesize that this may lead to similar results on training. More details on the training process are mentioned in Appendix \ref{sec:appendix}.

\section{Results \& Discussions}

We present the results of all three challenge sets separately.

\subsection{Set I: Temp-Order}

Results of \textit{Temp-Order} set are summarised in Table \ref{tab:Results CS1}. When evaluating on the RoBERTa and DeBERTa models, there is an improvement of about 10\% over the majority baseline. On analyzing the effect of different variations mentioned in Table \ref{tab:Variations in Temporal Ordering}, we find that changing the template position or its tense does not produce any significant difference in performance. However, we find that pairs where the premise event occurred at a fixed time instance (\ref{tab:Variations in Temporal Ordering} \textit{a}) have an average of 75\% accuracy, while the pairs where the premise event occurred over a time interval (\ref{tab:Variations in Temporal Ordering} \textit{d-f}) have an average accuracy of 28\%. This implies that models trained on MNLI have some basic understanding of temporal ordering, especially in determining whether a fixed time instance is present in another time interval. However, it gets difficult to reason about the ordering between two time intervals, where discerning the label is also not as straightforward.

\begin{figure*}
	\centering
	\includegraphics[width=\textwidth,height=\textheight,keepaspectratio]{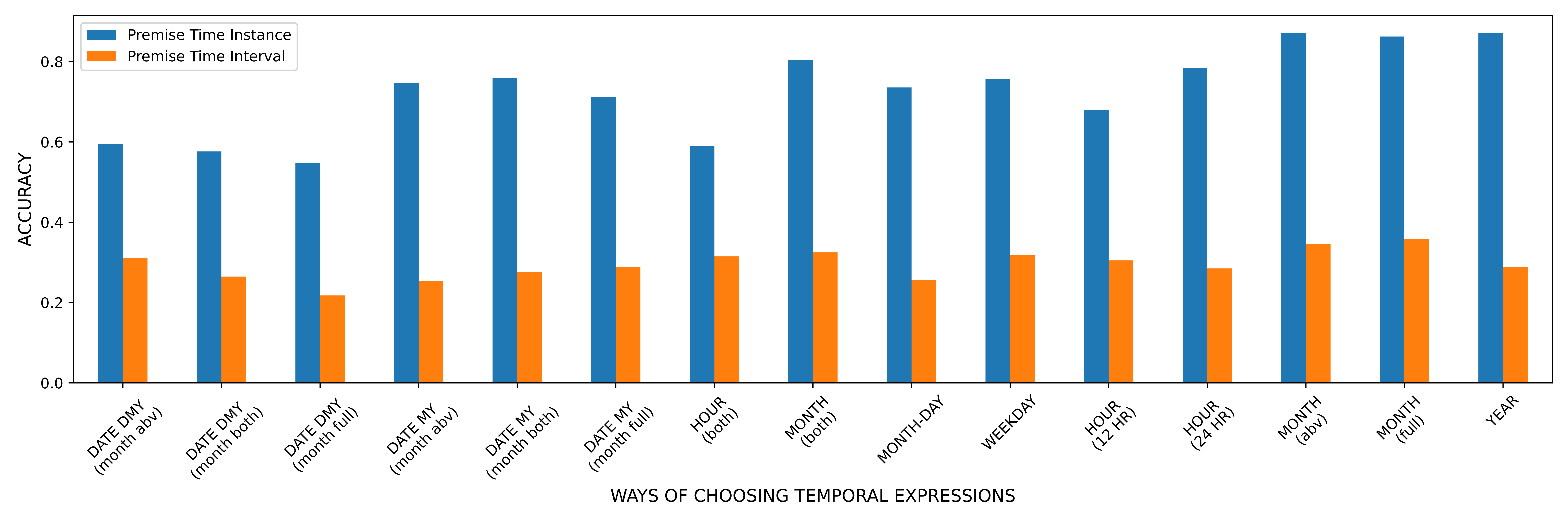}
	\rule{35em}{0.5pt}
	\caption{Comparison of accuracies across different ways of choosing temporal expressions when running \textit{Temp-Order} set on DeBERTa fine-tuned on MNLI. Label `DATE DMY' implies generating dates using all \textit{month-day}, \textit{month} and \textit{year} (Table \ref{tab:Variations in Temporal Ordering} \textit{j}), while `DATE MY' uses only \textit{month} and \textit{year} (Table \ref{tab:Variations in Temporal Ordering} \textit{i}). The added descriptions signify where the month instances are drawn from, where `month (full)' implies that both months come from the \textit{month (full name)} list, `month (abv)' implies that they come from \textit{month (abbreviated)} list, and `month (both)' implies that one month instance comes from each of the two lists. Labels `MONTH (both)' and `HOUR (both)' signify similar methods of choosing from multiple lists (Table \ref{tab:Variations in Temporal Ordering} \textit{h} and \textit{g} correspondingly).}
	\label{fig:CS1 MNLI Comparison}
\end{figure*}

We further analyze the accuracies for different methods of choosing the two time instances, and the results for DeBERTa are summed up in Figure \ref{fig:CS1 MNLI Comparison}. For pairs where the premise event takes place at a fixed point in time, most methods in which both time instances were sampled from the same list give over 73\% accuracy. Among the methods in which time instances are sampled across multiple lists, dates in which \textit{months} are combined with \textit{years} give an average accuracy of 74\%, but this drops to 57\% when \textit{month-days} are added, signifying that the comparison of specific dates becomes too complicated for the NLI model. The model also has a hard time mapping \textit{hours} between 12 and 24-hour format, giving only 59\% accuracy. For pairs where the premise takes place over a time interval, the accuracies of all methods are below 35\%. 

We also evaluate the RoBERTa model trained on \textit{UDS-NLI (order)} corpus on our challenge set. However, the average performance was only 56\%, even below the majority baseline for the corpus, not indicating any significant transfer of knowledge from their task of event-based temporal ordering.

The hypothesis-only baseline for both the RoBERTa/DeBERTa models is exactly the majority baseline, which is not surprising as the true label cannot be determined without knowing the premise time instance. Further fine-tuning the pre-trained MNLI models on our \textit{Temp-Order} set leads to an almost perfect accuracy of 99\%. This is despite having separate templates for the train and test split, and having time instances randomly sampled from different lists. Since our method of generating labels was automated and depended on the values of the time instances, we infer that the numerous parameters of a large language model were able to learn this label generation process from the artificial NLI data.

\subsection{Set II: Temp-Duration}

Results for \textit{Temp-Duration} set are summarized in Table \ref{tab:Results CS2}. Both the RoBERTa and DeBERTa models fine-tuned on MNLI produce poor accuracies of around 55\%, just over the majority baseline. While analyzing the DeBERTa predictions, we found that the model produced \textit{entailment} for 83\% of the data points, implying that it is not able to adequately determine durations. The different variations (Table \ref{tab:Variations in Temporal Duration}) or methods of sampling time instances also did not have any significant effect. We investigated the performances of the six types of hypotheses and found that among the hypothesis types with the \textit{contradiction} gold label, \textit{`equal to GOLD*10'} produced 0.68 F1-score, compared to \textit{`equal to GOLD+1'} and \textit{`less than GOLD'}, which produced 0.15 and 0.07 F1-scores respectively. This might indicate that while the NLI model has difficulty figuring out when the claimed duration is incorrect, it still does better off when it is very distant \textit{(GOLD*10)} from the gold duration compared to when it is very close \textit{(GOLD+1)}. We also find that for the \textit{`equal to GOLD'} hypothesis, pairs of instances far apart in a list tend to be misclassified more than pairs of instances that are closer to each other. This implies that determination of exact duration gets difficult for the NLI model as the distance between instances increases.

We also evaluate the model trained on \textit{UDS-NLI (duration)} corpus on our set, and the results are slightly above the majority baseline. While the predictions by this model were not as skewed, we could not find any significant impact of the variations or the different methods of sampling time instances, not indicating any possible knowledge transfer from their problem of determining coarse-grained event duration.

\begin{table*}
\centering
    \begin{tabular}{cccc}
    \hline
      \textbf{Model} & \textbf{Method} & \textbf{Accuracy} & \textbf{F1 Score} \\
      \hline
      \textbf{Majority}\footnotemark[1] & Binary Classification & 50.00 & 33.33 \\
      \hline
      \multirow{3}{*}{\textbf{RoBERTa (MNLI)}} & Direct Evaluation & 54.32 & 46.64 \\
      & Hypothesis Only Training & 68.82 $\pm$ 0.22 & 66.84 $\pm$ 0.51 \\
      & Train and Evaluate & 91.86 $\pm$ 6.42 & 91.85 $\pm$ 6.42 \\
      \hline
      \multirow{3}{*}{\textbf{DeBERTa (MNLI)}} & Direct Evaluation & 56.67 & 51.53 \\
      & Hypothesis Only Training & 64.15 $\pm$ 0.27 & 59.64 $\pm$ 0.41 \\
      & Train and Evaluate & 73.72 $\pm$ 3.92 & 73.28 $\pm$ 4.50 \\
      \hline
      \multirow{1}{*}{\textbf{UDS-NLI (duration)}} & Direct Evaluation & 58.44 & 57.66 \\
      \hline
    \end{tabular}
    \caption{Evaluating \textit{Temp-Duration} set on NLI models}
    \label{tab:Results CS2}
\end{table*}

\begin{table*}
\centering
    \begin{tabular}{cccc}
      \hline
      \textbf{Model} & \textbf{Method} & \textbf{Accuracy} & \textbf{F1 Score} \\
      \hline
      \multirow{2}{*}{\textbf{Majority}} & Ternary Classification & 33.33 & 16.67 \\
      & Binary Classification & 66.67 & 53.33 \\
      \hline
      \multirow{3}{*}{\textbf{RoBERTa (MNLI)}} & Direct Evaluation & 35.47 & 28.71 \\
      & Hypothesis Only Training & 49.38 $\pm$ 0.71 & 39.51 $\pm$ 0.52 \\
      & Train and Evaluate & 99.97 $\pm$ 0.02 & 99.97 $\pm$ 0.02 \\
      \hline
      \multirow{3}{*}{\textbf{DeBERTa (MNLI)}} & Direct Evaluation & 45.02 & 38.60 \\
      & Hypothesis Only Training & 49.58 $\pm$ 0.79 & 41.29 $\pm$ 0.56 \\
      & Train and Evaluate & 99.94 $\pm$ 0.03 & 99.94 $\pm$ 0.03 \\
      \hline
      \multirow{1}{*}{\textbf{UDS-NLI (duration)}} & Direct Evaluation & 52.61 & 53.54 \\
      \hline
    \end{tabular}
    \caption{Evaluating \textit{Cross-Unit Duration} set on NLI models}
    \label{tab:Results CS3}
\end{table*}

On fine-tuning our challenge set under the hypothesis-only setting, both MNLI models surprisingly produce at least 15\% gains in accuracy over the majority baseline. We investigate and find that the models use lexical cues from the different hypothesis types, producing \textit{entailment} for all \textit{`less than'} hypotheses. For the \textit{`equal'} hypotheses, they predict \textit{contradiction} when the claimed duration is a large value (more likely to be \textit{GOLD*10}) and \textit{entailment} when it is smaller (more likely to be \textit{GOLD}). However, under the standard NLI training scenario, these cues are not the only factor behind learning, as the RoBERTa MNLI model produces 91.86\% average accuracy, which is a gain of 20\% over the hypothesis-only setting. Among the various methods of sampling time instances, \textit{years} performs the worst, producing only 66\% accuracy, possibly because the lengths of duration can be as large as 100 years. Finally, the hypothesis types \textit{`equal to GOLD*10'} and \textit{`less than GOLD*10'} produce 99\% accuracy, while \textit{`equal to GOLD'} and \textit{`less than GOLD'} report below 90\%, confirming our speculation that it is easier for the models to reason about the validity of the claimed duration when it is distant from the gold duration.

\subsection{Set III: Cross-Unit Duration Set}

As shown in Table \ref{tab:Results CS3}, all the models produce poor performances on direct evaluation, just near the majority baseline. DeBERTa fine-tuned on MNLI manages to perform better when compared to RoBERTa fine-tuned on MNLI by around 10\% on overall accuracy. Hence, we can conclude that DeBERTa has a slightly better understanding of cross-unit duration comparison when compared to RoBERTa.

\footnotetext[1]{Same majority baseline because no neutral labels.}

Similar to \textit{Temp-Order} set, all models performed better compared to their respective majority baselines when the premise event occurs at a future point in time rather than over a time interval. More specifically, we see an improvement in accuracy of around 18\% (29.65 to 47.12) for RoBERTa and around 10\% (41.77 to 51.52) for DeBERTa when we switch from the premise occurring over a time interval to a point in time.

We analyzed the results on adjacent units to recognize if there are specific pairs for which the models are better able to figure out the conversion relationship. We did not find any significant pair for RoBERTa or DeBERTa models, but we find that the \textit{UDS-NLI (duration)} model does better on bigger unit pairs of duration, i.e., it performs the best on conversion between month-years (56.03\% F1), then day-months (55.32\% F1), then hours-days (51.02\% F1), and then minutes-hours (16.74\% F1). This suggests a better transfer of knowledge for bigger time units from the \textit{UDS-NLI (duration)} model to our challenge set.

Similar to \textit{Temp-Duration} set, on fine-tuning under the hypothesis-only setting, both MNLI models produce around 16-17\% gains in accuracy over the majority baseline using lexical cues present in the hypotheses due to the challenge set creation process. For the hypotheses that contain \textit{`before'}, the models tend to predict \textit{entailment} if the duration \textit{(T2)} is large, and \textit{contradiction} if it is small. Similarly, for the hypotheses that contain \textit{`after'}, the models mostly predict \textit{contradiction} if the duration is large, and \textit{entailment} if it is small. However, on standard training, the accuracy goes up from around 50\% to near perfect 99\%, showing that these cues are not the only reason behind the model's performance and that it actually learns the relationship between the premise and hypothesis.

We believe a valuable addition to this challenge set would be introducing more varied phrasing of prepositions. That is, using synonymous ways of denoting a temporal event occurring before, after, or strictly at a point in time. In particular, phrasing like `after the next 60 minutes' or `sometime after 60 minutes pass' could be examples of more specific ways to represent that an event occurred `after 60 minutes' - a phrase which we acknowledge may read to mean `in exactly 60 minutes', as opposed to some time after.

\section{Conclusion}

We create three challenge sets that test different kinds of relationships between temporal expressions. We evaluate these challenge sets on popular NLI models like RoBERTa and DeBERTa trained on MNLI, and find that while they can reason about simple cases of ordering between time instances, they fail when presented with more complicated cases or when temporal reasoning requires determining fine-grained duration. Since our challenge sets were synthetically created, training on them helped the NLI models to figure out the label generation process, and they produced near-perfect accuracy for the \textit{Temp-Order} and the \textit{Cross-Unit Duration} sets. A direction for future research could be evaluating and comparing models, trained on other NLI datasets containing temporal expressions, on our challenge sets. Another direction could be to collect naturally occurring sentences that contain temporal expressions from large corpora and recast them into NLI format for similar testing of understanding of temporal expressions. 

\bibliography{anthology,custom}

\begin{thebibliography}{28}
\expandafter\ifx\csname natexlab\endcsname\relax\def\natexlab#1{#1}\fi

\bibitem[{Bethard et~al.(2015)Bethard, Derczynski, Savova, Pustejovsky, and
  Verhagen}]{bethard-etal-2015-semeval}
Steven Bethard, Leon Derczynski, Guergana Savova, James Pustejovsky, and Marc
  Verhagen. 2015.
\newblock \href {https://doi.org/10.18653/v1/S15-2136} {{S}em{E}val-2015 task
  6: Clinical {T}emp{E}val}.
\newblock In \emph{Proceedings of the 9th International Workshop on Semantic
  Evaluation ({S}em{E}val 2015)}, pages 806--814, Denver, Colorado. Association
  for Computational Linguistics.

\bibitem[{Cassidy et~al.(2014)Cassidy, McDowell, Chambers, and
  Bethard}]{cassidy-etal-2014-annotation}
Taylor Cassidy, Bill McDowell, Nathanael Chambers, and Steven Bethard. 2014.
\newblock \href {https://doi.org/10.3115/v1/P14-2082} {An annotation framework
  for dense event ordering}.
\newblock In \emph{Proceedings of the 52nd Annual Meeting of the Association
  for Computational Linguistics (Volume 2: Short Papers)}, pages 501--506,
  Baltimore, Maryland. Association for Computational Linguistics.

\bibitem[{Chen et~al.(2019)Chen, Wang, and
  Karlsson}]{timeexpressionrecognition}
Sanxing Chen, Guoxin Wang, and B{\"o}rje Karlsson. 2019.
\newblock \href {https://www.cs.virginia.edu/~sc3hn/bert-time.pdf} {Exploring
  word representations on time expression recognition}.
\newblock Technical report, Microsoft Research Asia.

\bibitem[{Devlin et~al.(2019)Devlin, Chang, Lee, and
  Toutanova}]{devlin-etal-2019-bert}
Jacob Devlin, Ming-Wei Chang, Kenton Lee, and Kristina Toutanova. 2019.
\newblock \href {https://doi.org/10.18653/v1/N19-1423} {{BERT}: Pre-training of
  deep bidirectional transformers for language understanding}.
\newblock In \emph{Proceedings of the 2019 Conference of the North {A}merican
  Chapter of the Association for Computational Linguistics: Human Language
  Technologies, Volume 1 (Long and Short Papers)}, pages 4171--4186,
  Minneapolis, Minnesota. Association for Computational Linguistics.

\bibitem[{Ding et~al.(2019)Ding, Gao, Shi, and Qu}]{Ding_Gao_Shi_Qu_2019}
Wentao Ding, Guanji Gao, Linfeng Shi, and Yuzhong Qu. 2019.
\newblock \href {https://doi.org/10.1609/aaai.v33i01.33016335} {A pattern-based
  approach to recognizing time expressions}.
\newblock \emph{Proceedings of the AAAI Conference on Artificial Intelligence},
  33(01):6335--6342.

\bibitem[{Do et~al.(2012)Do, Lu, and Roth}]{do-etal-2012-joint}
Quang Do, Wei Lu, and Dan Roth. 2012.
\newblock \href {https://aclanthology.org/D12-1062} {Joint inference for event
  timeline construction}.
\newblock In \emph{Proceedings of the 2012 Joint Conference on Empirical
  Methods in Natural Language Processing and Computational Natural Language
  Learning}, pages 677--687, Jeju Island, Korea. Association for Computational
  Linguistics.

\bibitem[{He et~al.(2021)He, Liu, Gao, and Chen}]{he2021deberta}
Pengcheng He, Xiaodong Liu, Jianfeng Gao, and Weizhu Chen. 2021.
\newblock \href {http://arxiv.org/abs/2006.03654} {Deberta: Decoding-enhanced
  bert with disentangled attention}.

\bibitem[{Lee et~al.(2014)Lee, Artzi, Dodge, and
  Zettlemoyer}]{lee-etal-2014-context}
Kenton Lee, Yoav Artzi, Jesse Dodge, and Luke Zettlemoyer. 2014.
\newblock \href {https://doi.org/10.3115/v1/P14-1135} {Context-dependent
  semantic parsing for time expressions}.
\newblock In \emph{Proceedings of the 52nd Annual Meeting of the Association
  for Computational Linguistics (Volume 1: Long Papers)}, pages 1437--1447,
  Baltimore, Maryland. Association for Computational Linguistics.

\bibitem[{Leeuwenberg and Moens(2018)}]{leeuwenberg-moens-2018-temporal}
Artuur Leeuwenberg and Marie-Francine Moens. 2018.
\newblock \href {https://doi.org/10.18653/v1/D18-1155} {Temporal information
  extraction by predicting relative time-lines}.
\newblock In \emph{Proceedings of the 2018 Conference on Empirical Methods in
  Natural Language Processing}, pages 1237--1246, Brussels, Belgium.
  Association for Computational Linguistics.

\bibitem[{Liu et~al.(2019)Liu, Ott, Goyal, Du, Joshi, Chen, Levy, Lewis,
  Zettlemoyer, and Stoyanov}]{liu2019roberta}
Yinhan Liu, Myle Ott, Naman Goyal, Jingfei Du, Mandar Joshi, Danqi Chen, Omer
  Levy, Mike Lewis, Luke Zettlemoyer, and Veselin Stoyanov. 2019.
\newblock \href {http://arxiv.org/abs/1907.11692} {Roberta: A robustly
  optimized bert pretraining approach}.

\bibitem[{McCoy et~al.(2019)McCoy, Pavlick, and Linzen}]{mccoy-etal-2019-right}
Tom McCoy, Ellie Pavlick, and Tal Linzen. 2019.
\newblock \href {https://doi.org/10.18653/v1/P19-1334} {Right for the wrong
  reasons: Diagnosing syntactic heuristics in natural language inference}.
\newblock In \emph{Proceedings of the 57th Annual Meeting of the Association
  for Computational Linguistics}, pages 3428--3448, Florence, Italy.
  Association for Computational Linguistics.

\bibitem[{Naik et~al.(2019)Naik, Breitfeller, and
  Rose}]{naik-etal-2019-tddiscourse}
Aakanksha Naik, Luke Breitfeller, and Carolyn Rose. 2019.
\newblock \href {https://doi.org/10.18653/v1/W19-5929} {{TDD}iscourse: A
  dataset for discourse-level temporal ordering of events}.
\newblock In \emph{Proceedings of the 20th Annual SIGdial Meeting on Discourse
  and Dialogue}, pages 239--249, Stockholm, Sweden. Association for
  Computational Linguistics.

\bibitem[{Ning et~al.(2020)Ning, Wu, Han, Peng, Gardner, and
  Roth}]{ning2020torque}
Qiang Ning, Hao Wu, Rujun Han, Nanyun Peng, Matt Gardner, and Dan Roth. 2020.
\newblock \href {http://arxiv.org/abs/2005.00242} {Torque: A reading
  comprehension dataset of temporal ordering questions}.

\bibitem[{Ning et~al.(2018)Ning, Zhou, Feng, Peng, and
  Roth}]{ning-etal-2018-cogcomptime}
Qiang Ning, Ben Zhou, Zhili Feng, Haoruo Peng, and Dan Roth. 2018.
\newblock \href {https://doi.org/10.18653/v1/D18-2013} {{C}og{C}omp{T}ime: A
  tool for understanding time in natural language}.
\newblock In \emph{Proceedings of the 2018 Conference on Empirical Methods in
  Natural Language Processing: System Demonstrations}, pages 72--77, Brussels,
  Belgium. Association for Computational Linguistics.

\bibitem[{Poliak et~al.(2018)Poliak, Naradowsky, Haldar, Rudinger, and
  Van~Durme}]{poliak-etal-2018-hypothesis}
Adam Poliak, Jason Naradowsky, Aparajita Haldar, Rachel Rudinger, and Benjamin
  Van~Durme. 2018.
\newblock \href {https://doi.org/10.18653/v1/S18-2023} {Hypothesis only
  baselines in natural language inference}.
\newblock In \emph{Proceedings of the Seventh Joint Conference on Lexical and
  Computational Semantics}, pages 180--191, New Orleans, Louisiana. Association
  for Computational Linguistics.

\bibitem[{Pustejovsky et~al.(2003{\natexlab{a}})Pustejovsky, Castano, Ingria,
  Sauri, Gaizauskas, Setzer, Katz, and Radev}]{pustejovsky2003timeml}
James Pustejovsky, Jos{\'e}~M Castano, Robert Ingria, Roser Sauri, Robert~J
  Gaizauskas, Andrea Setzer, Graham Katz, and Dragomir~R Radev.
  2003{\natexlab{a}}.
\newblock Timeml: Robust specification of event and temporal expressions in
  text.
\newblock \emph{New directions in question answering}, 3:28--34.

\bibitem[{Pustejovsky et~al.(2003{\natexlab{b}})Pustejovsky, Hanks, Saurí,
  See, Gaizauskas, Setzer, Radev, Sundheim, Day, Ferro, and
  Lazo}]{pustejovsky2003timebank}
James Pustejovsky, Patrick Hanks, Roser Saurí, Andrew See, Rob Gaizauskas,
  Andrea Setzer, Dragomir Radev, Beth Sundheim, David Day, Lisa Ferro, and
  Marcia Lazo. 2003{\natexlab{b}}.
\newblock The timebank corpus.
\newblock \emph{Proceedings of Corpus Linguistics}, page~40.

\bibitem[{Rogers et~al.(2020)Rogers, Kovaleva, and
  Rumshisky}]{rogers-etal-2020-primer}
Anna Rogers, Olga Kovaleva, and Anna Rumshisky. 2020.
\newblock \href {https://doi.org/10.1162/tacl_a_00349} {A primer in
  {BERT}ology: What we know about how {BERT} works}.
\newblock \emph{Transactions of the Association for Computational Linguistics},
  8:842--866.

\bibitem[{Str{\"o}tgen and Gertz(2010)}]{strotgen-gertz-2010-heideltime}
Jannik Str{\"o}tgen and Michael Gertz. 2010.
\newblock \href {https://aclanthology.org/S10-1071} {{H}eidel{T}ime: High
  quality rule-based extraction and normalization of temporal expressions}.
\newblock In \emph{Proceedings of the 5th International Workshop on Semantic
  Evaluation}, pages 321--324, Uppsala, Sweden. Association for Computational
  Linguistics.

\bibitem[{UzZaman et~al.(2013)UzZaman, Llorens, Derczynski, Allen, Verhagen,
  and Pustejovsky}]{uzzaman-etal-2013-semeval}
Naushad UzZaman, Hector Llorens, Leon Derczynski, James Allen, Marc Verhagen,
  and James Pustejovsky. 2013.
\newblock \href {https://aclanthology.org/S13-2001} {{S}em{E}val-2013 task 1:
  {T}emp{E}val-3: Evaluating time expressions, events, and temporal relations}.
\newblock In \emph{Second Joint Conference on Lexical and Computational
  Semantics (*{SEM}), Volume 2: Proceedings of the Seventh International
  Workshop on Semantic Evaluation ({S}em{E}val 2013)}, pages 1--9, Atlanta,
  Georgia, USA. Association for Computational Linguistics.

\bibitem[{Vashishtha et~al.(2020)Vashishtha, Poliak, Lal, Van~Durme, and
  White}]{vashishtha-etal-2020-temporal}
Siddharth Vashishtha, Adam Poliak, Yash~Kumar Lal, Benjamin Van~Durme, and
  Aaron~Steven White. 2020.
\newblock \href {https://doi.org/10.18653/v1/2020.findings-emnlp.363} {Temporal
  reasoning in natural language inference}.
\newblock In \emph{Findings of the Association for Computational Linguistics:
  EMNLP 2020}, pages 4070--4078, Online. Association for Computational
  Linguistics.

\bibitem[{Vashishtha et~al.(2019)Vashishtha, Van~Durme, and
  White}]{vashishtha-etal-2019-fine}
Siddharth Vashishtha, Benjamin Van~Durme, and Aaron~Steven White. 2019.
\newblock \href {https://doi.org/10.18653/v1/P19-1280} {Fine-grained temporal
  relation extraction}.
\newblock In \emph{Proceedings of the 57th Annual Meeting of the Association
  for Computational Linguistics}, pages 2906--2919, Florence, Italy.
  Association for Computational Linguistics.

\bibitem[{Verhagen et~al.(2007)Verhagen, Gaizauskas, Schilder, Hepple, Katz,
  and Pustejovsky}]{verhagen-etal-2007-semeval}
Marc Verhagen, Robert Gaizauskas, Frank Schilder, Mark Hepple, Graham Katz, and
  James Pustejovsky. 2007.
\newblock \href {https://aclanthology.org/S07-1014} {{S}em{E}val-2007 task 15:
  {T}emp{E}val temporal relation identification}.
\newblock In \emph{Proceedings of the Fourth International Workshop on Semantic
  Evaluations ({S}em{E}val-2007)}, pages 75--80, Prague, Czech Republic.
  Association for Computational Linguistics.

\bibitem[{Verhagen et~al.(2010)Verhagen, Saur{\'\i}, Caselli, and
  Pustejovsky}]{verhagen-etal-2010-semeval}
Marc Verhagen, Roser Saur{\'\i}, Tommaso Caselli, and James Pustejovsky. 2010.
\newblock \href {https://aclanthology.org/S10-1010} {{S}em{E}val-2010 task 13:
  {T}emp{E}val-2}.
\newblock In \emph{Proceedings of the 5th International Workshop on Semantic
  Evaluation}, pages 57--62, Uppsala, Sweden. Association for Computational
  Linguistics.

\bibitem[{Wallace et~al.(2019)Wallace, Wang, Li, Singh, and
  Gardner}]{wallace-etal-2019-nlp}
Eric Wallace, Yizhong Wang, Sujian Li, Sameer Singh, and Matt Gardner. 2019.
\newblock \href {https://doi.org/10.18653/v1/D19-1534} {Do {NLP} models know
  numbers? probing numeracy in embeddings}.
\newblock In \emph{Proceedings of the 2019 Conference on Empirical Methods in
  Natural Language Processing and the 9th International Joint Conference on
  Natural Language Processing (EMNLP-IJCNLP)}, pages 5307--5315, Hong Kong,
  China. Association for Computational Linguistics.

\bibitem[{Williams et~al.(2018)Williams, Nangia, and
  Bowman}]{williams-etal-2018-broad}
Adina Williams, Nikita Nangia, and Samuel Bowman. 2018.
\newblock \href {https://doi.org/10.18653/v1/N18-1101} {A broad-coverage
  challenge corpus for sentence understanding through inference}.
\newblock In \emph{Proceedings of the 2018 Conference of the North {A}merican
  Chapter of the Association for Computational Linguistics: Human Language
  Technologies, Volume 1 (Long Papers)}, pages 1112--1122, New Orleans,
  Louisiana. Association for Computational Linguistics.

\bibitem[{Wolf et~al.(2020)Wolf, Debut, Sanh, Chaumond, Delangue, Moi, Cistac,
  Rault, Louf, Funtowicz, Davison, Shleifer, von Platen, Ma, Jernite, Plu, Xu,
  Scao, Gugger, Drame, Lhoest, and Rush}]{wolf2020huggingfaces}
Thomas Wolf, Lysandre Debut, Victor Sanh, Julien Chaumond, Clement Delangue,
  Anthony Moi, Pierric Cistac, Tim Rault, Rémi Louf, Morgan Funtowicz, Joe
  Davison, Sam Shleifer, Patrick von Platen, Clara Ma, Yacine Jernite, Julien
  Plu, Canwen Xu, Teven~Le Scao, Sylvain Gugger, Mariama Drame, Quentin Lhoest,
  and Alexander~M. Rush. 2020.
\newblock \href {http://arxiv.org/abs/1910.03771} {Huggingface's transformers:
  State-of-the-art natural language processing}.

\bibitem[{Zhou et~al.(2019)Zhou, Khashabi, Ning, and
  Roth}]{zhou-etal-2019-going}
Ben Zhou, Daniel Khashabi, Qiang Ning, and Dan Roth. 2019.
\newblock \href {https://doi.org/10.18653/v1/D19-1332} {{``}going on a
  vacation{''} takes longer than {``}going for a walk{''}: A study of temporal
  commonsense understanding}.
\newblock In \emph{Proceedings of the 2019 Conference on Empirical Methods in
  Natural Language Processing and the 9th International Joint Conference on
  Natural Language Processing (EMNLP-IJCNLP)}, pages 3363--3369, Hong Kong,
  China. Association for Computational Linguistics.

\end{thebibliography}
\bibliographystyle{acl_natbib}

\appendix

\section{Training Details}
\label{sec:appendix}

We use three different NLI models for our experiments. For the RoBERTa and DeBERTa models fine-tuned on MNLI, we use the \texttt{roberta-large-mnli} and \texttt{microsoft/deberta-large-mnli} models respectively, available under the \texttt{transformers} library from HuggingFace \citep{wolf2020huggingfaces}. For the \textit{UDS-NLI} models, we directly use the saved RoBERTa-large models for \textit{UDS-NLI (duration)} and \textit{UDS-NLI (order)} made publicly available by \citet{vashishtha-etal-2020-temporal}.

For training, we use an Adam optimizer, with a learning rate of $2*10^{-5}$ and $0.1$ weight decay. We use a batch size of 16 for training and 128 for testing. We train for 10 epochs, using early stopping with a patience of 2. We run each experiment for three random seeds (3, 5, 7), and use them to calculate the mean and standard deviation for the accuracy and F1 (weighted) score metrics.

\end{document}